%% file: main.tex
\documentclass[10pt,twocolumn,letterpaper]{article}

\usepackage[pagenumbers]{cvpr}
\input{preamble}

\definecolor{cvprblue}{rgb}{0.21,0.49,0.74}
\usepackage[pagebackref,breaklinks,colorlinks,citecolor=cvprblue]{hyperref}

\def\paperID{3710} \def\confName{CVPR}
\def\confYear{2024}

\title{Revisiting Non-Autoregressive Transformers for Efficient Image Synthesis}

\author{
Zanlin Ni$^{1*}$\ \ \ \
Yulin Wang$^{1}$\thanks{Equal contribution.} \ \ \
Renping Zhou$^{1}$ \ \ \
Jiayi Guo$^{1}$\\
Jinyi Hu$^{1}$ \ \ \
Zhiyuan Liu$^{1}$ \ \ \
Shiji Song$^{1}$ \ \ \
Yuan Yao$^{2\dagger}$ \ \ \
Gao Huang$^{1}$\thanks{Corresponding authors.} \\
{\small $^{1}$Tsinghua University \ \ \ $^{2}$National University of Singapore} \\
}

\begin{document}

\maketitle
\setlength\abovedisplayskip{4pt}
\setlength\belowdisplayskip{4pt}
\input{sec/0_abstract}
\input{sec/1_intro}

\input{sec/2_related.tex}
\input{sec/3_1_preliminaries}

\input{sec/3_3_method}

\input{sec/4_experiment.tex}

\input{sec/conclusion.tex}
\input{sec/acknowledgements.tex}

{\small
\bibliographystyle{ieeenat_fullname}
\bibliography{main}
}

\end{document}

%% file: preamble.tex
\usepackage[dvipsnames]{xcolor}

\usepackage{graphicx}
\usepackage{amsmath}
\DeclareMathOperator*{\argmax}{arg\,max}
\usepackage{amssymb}
\usepackage{booktabs}

\usepackage{bm}
\usepackage{algorithm}
\usepackage{algpseudocode} \usepackage{multirow}
\usepackage{wrapfig}
\usepackage{amsthm}
\usepackage{caption}
\usepackage{listings}
\usepackage{algpseudocode}\usepackage[T1]{fontenc}
\usepackage{xcolor}
\usepackage{colortbl}

\newlength\savewidth\newcommand\shline{\noalign{\global\savewidth\arrayrulewidth
\global\arrayrulewidth 1pt}\hline\noalign{\global\arrayrulewidth\savewidth}}

\usepackage{array}
\renewcommand{\paragraph}[1]{\noindent\textbf{#1}}

\newcommand{\tablestyle}[2]{\setlength{\tabcolsep}{#1}\renewcommand{\arraystretch}{#2}\centering\footnotesize}

\definecolor{defaultcolor}{gray}{.9}
\newcommand{\default}[1]{\cellcolor{defaultcolor}{#1}}
\definecolor{ourscolor}{HTML}{EFFAF1}

\definecolor{deemph}{gray}{0.6}
\newcommand{\gc}[1]{\textcolor{deemph}{#1}}
\definecolor{decoded_color}{HTML}{A5DCE0}
\definecolor{masked_color}{HTML}{A7A9AC}

\definecolor{ratio}{rgb}{0.0, 0.0, 0.0}
\definecolor{sampt}{rgb}{0.0, 0.0, 0.0}
\definecolor{remaskt}{rgb}{0.0, 0.0, 0.0}
\definecolor{cfg}{rgb}{0.0, 0.0, 0.0}

\usepackage{pifont}
\newcommand{\cmark}{\ding{51}}

\usepackage{lipsum}

\newcommand{\pub}[1]{\xspace\textcolor{gray}{\tiny{(\textit{#1})}}}
\definecolor{Highlight}{HTML}{39b54a}  \newcommand{\hl}[1]{\textcolor{Highlight}{\textbf{#1}}}
\newcommand{\dt}[2]{\textbf{#1}\fontsize{5.5pt}{0.1em}\selectfont~\hl{(#2)}}

\usepackage{xspace}
\newcommand{\VQVAE}{VQVAE-2$^\ddagger$~\cite{razavi2019generating}\pub{NeurIPS'19}\xspace}

\newcommand{\LDM}{LDM~\cite{rombach2022high}\pub{CVPR'22}\xspace}

\newcommand{\ADMp}{ADM-G$^\dagger$~\cite{dhariwal2021diffusion}\pub{NeurIPS'21}\xspace}
\newcommand{\LDMp}{LDM$^\dagger$~\cite{rombach2022high}\pub{CVPR'22}\xspace}
\newcommand{\UVITp}{U-ViT-H$^\dagger$~\cite{bao2022all}\pub{CVPR'23}\xspace}
\newcommand{\DITp}{DiT-XL$^\dagger$~\cite{peebles2023scalable}\pub{ICCV'23}\xspace}

\newcommand{\VQGAN}{VQGAN$^\ddagger$~\cite{esser2021taming}\pub{CVPR'21}\xspace}
\newcommand{\VQDiffusion}{VQ-Diffusion$^\ddagger$~\cite{gu2022vector}\pub{CVPR'22}\xspace}
\newcommand{\MaskGIT}{MaskGIT$^\ddagger$~\cite{chang2022maskgit}\pub{CVPR'22}\xspace}

\newcommand{\TokenCritic}{Token-Critic$^\ddagger$~\cite{lezama2022improved}\pub{ECCV'22}\xspace}
\newcommand{\DraftRevise}{Draft-and-revise~\cite{lee2022draft}\pub{NeurIPS'22}\xspace}
\newcommand{\MAGE}{MAGE$^\ddagger$~\cite{li2023mage}\pub{CVPR'23}\xspace}
\newcommand{\Ours}{AutoNAT\xspace}
\newcommand{\Solver}{DPM-Solver~\cite{lu2022dpmp}\pub{NeurIPS'22}\xspace}

\newcolumntype{x}[1]{>{\centering\arraybackslash}p{#1pt}}
\newcolumntype{y}[1]{>{\raggedright\arraybackslash}p{#1pt}}
\newcolumntype{z}[1]{>{\raggedleft\arraybackslash}p{#1pt}}

%% file: sec/0_abstract.tex
\begin{abstract}
    The field of image synthesis is currently flourishing due to the advancements in diffusion models.
    While diffusion models have been successful, their computational intensity has prompted the pursuit of more efficient alternatives.
    As a representative work, non-autoregressive Transformers (NATs) have been recognized for their rapid generation.
    However, a major drawback of these models is their inferior performance compared to diffusion models.
    In this paper, we aim to re-evaluate the full potential of NATs by revisiting the design of their training and inference strategies.
    Specifically, we identify the complexities in properly configuring these strategies and indicate the possible sub-optimality in existing heuristic-driven designs.
    Recognizing this, we propose to go beyond existing methods by directly solving the optimal strategies in an automatic framework.
    The resulting method, named \Ours, advances the performance boundaries of NATs notably, and is able to perform comparably with the latest diffusion models at a significantly reduced inference cost.
    The effectiveness of \Ours is validated on four benchmark datasets, i.e., ImageNet-256 \& 512, MS-COCO, and CC3M.
    Our code is available at \url{https://github.com/LeapLabTHU/ImprovedNAT}.
\end{abstract}

%% file: sec/1_intro.tex
\section{Introduction}
\label{sec:intro}
In recent years, the field of image synthesis has witnessed a surge in popularity due to the success of highly capable diffusion models~\cite{ho2020denoising,song2020denoising,dhariwal2021diffusion,rombach2022high,ramesh2022hierarchical}.
However, the iterative generation process of diffusion models tends to be computationally intensive~\cite{song2020denoising,esser2021taming,yu2022scaling,lu2022dpm,lu2022dpmp}, which leads to significant practical latency and energy consumption.
As a consequence, a number of research efforts have been put into exploring more efficient alternatives.

\begin{figure}[!t]\centering
\includegraphics[width=\linewidth]{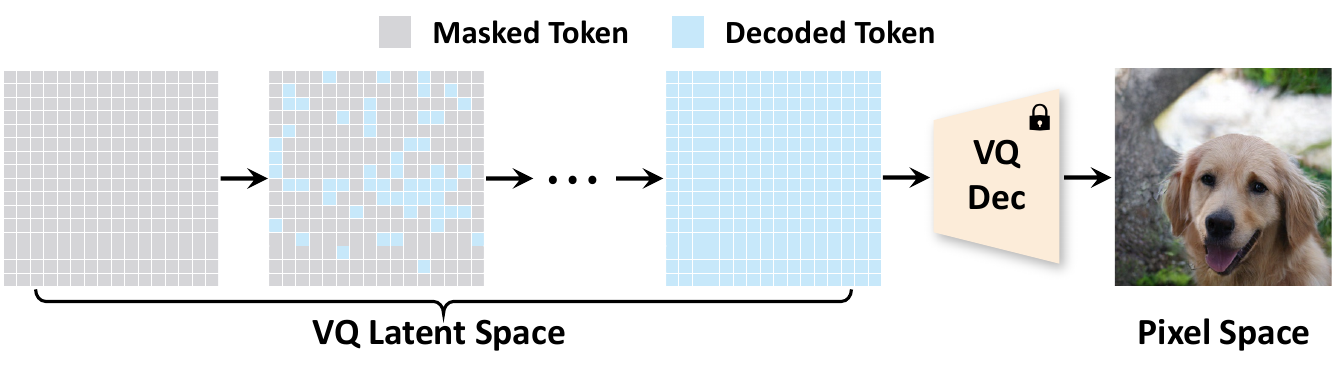}
\vskip -0.1in
\caption{
    \textbf{The generation process of non-autoregressive Transformers} starts from an entirely masked canvas and parallelly decodes multiple tokens at each step.
    The generated tokens are then mapped to the pixel space with a pre-trained VQ-decoder~\cite{esser2021taming}.
    \label{fig:illustrate_nar}
}
%\vskip -0.1in
\end{figure}

Within this context, non-autoregressive Transformers (NATs)~\cite{chang2022maskgit,lezama2022improved,chang2023muse,li2023mage,qian2023strait} have emerged as a representative work.
These models offer benefits akin to those of diffusion models, such as high scalability~\cite{chang2023muse} and sample diversity~\cite{chang2022maskgit}, while being notably faster due to their parallel decoding mechanism in Vector Quantized (VQ) space~\cite{van2017neural,esser2021taming}.
As depicted in Figure~\ref{fig:illustrate_nar}, NATs start generation from a fully masked canvas, and swiftly progress by concurrently decoding multiple tokens at each step.
Nevertheless, a major drawback of NATs is their inferior performance compared to diffusion models.
For example, MaskGIT~\cite{chang2022maskgit}, a popular NAT model, can synthesize an image on ImageNet~\cite{russakovsky2015imagenet} in only 8 steps, but the generation quality measured in FID is only 6.18, which is far behind latest diffusion models~\cite{peebles2023scalable,bao2022all} (approximately 2).

\begin{figure}[t]\centering
\includegraphics[width=.9\linewidth]{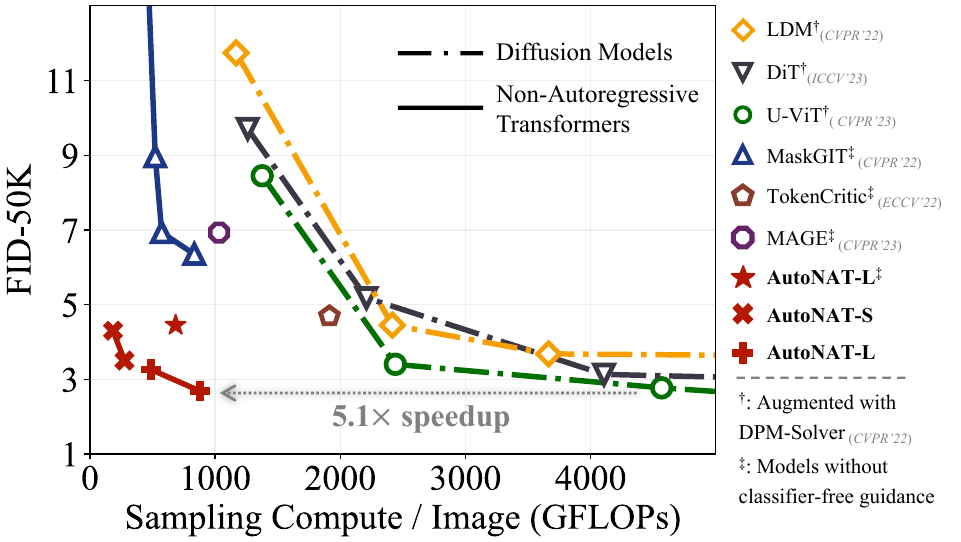}
%\vskip -0.1in
\caption{
    \textbf{FID-50K \vs computational cost on ImageNet-256}.
    For fair comparisons, diffusion models are equipped with DPM-Solver~\cite{lu2022dpm,lu2022dpmp} for efficient synthesis.
    \label{fig:tflops_imnet}
}
%\vskip -0.2in
\end{figure}

In this paper, we re-examine the performance limit of NATs.
Our findings demonstrate that the inferior generation quality compared to diffusion models may not be their inherent limitation.
Instead, it is largely caused by the sub-optimal, heuristic-driven strategies in the training and generation process of NATs.
To be specific, different from diffusion models, the parallel decoding mechanism in NATs introduces intricate design challenges, posing questions like 1) how many tokens should be decoded at each step; 2) which tokens should be decoded; and 3) how to sample tokens from the VQ codebook?
Configuring these aspects appropriately necessitates a specialized "\emph{generation strategy}" comprising multiple scheduling functions.
Moreover, a "\emph{training strategy}" needs to be carefully designed to equip the model with the ability to handle the varying input distributions encountered during generation.
Developing proper generation and training strategies that maximally unleash the potential of NATs is a critical but under-explored open problem.
The common practice in the literature predominantly designs these strategies with heuristic-driven rules (see: Tab.~\ref{tab:teaser})~\cite{chang2022maskgit,chang2023muse}.
This demands extensive expert knowledge and labor-intensive efforts, yet it can still result in sub-optimal configurations.

Contrary to existing practices, we introduce a heuristic-free, automatic approach, termed \Ours.
The major insight behind our method is to formulate the task of designing effective training and generation strategies into a unified optimization problem, as shown in Table~\ref{tab:teaser}.
This allows for a more comprehensive exploration of NATs' full potential without being constrained by the limited prior knowledge.
However, a notable challenge arises when optimizing the training strategy, as it involves a time-consuming model training process.
This requirement leads to a bottleneck in the optimization procedure, hindering the swift advancement of other optimization variables.
To address this challenge, we introduce a tailored alternating optimization algorithm, where the training and generation strategies are optimized alternatively according to their own characteristics.
This mechanism allows us to design specialized solutions conditioned on each individual sub-problem, resulting in a more efficient and focused optimization process.

The effectiveness of \Ours{} is extensively validated on four benchmark datasets, \emph{i.e.}, ImageNet-256 and 512~\cite{russakovsky2015imagenet}, MS-COCO~\cite{lin2014microsoft}, and CC3M~\cite{sharma2018conceptual}.
The results show that \Ours{} outperforms previous NATs by large margins.
Furthermore, compared to the diffusion models equipped with fast samplers, \Ours{} achieves at about $5\times$ inference speedup without sacrificing performance.

\input{tables/teaser.tex}

%% file: tables/teaser.tex
\begin{table}[t]
    \centering
    \resizebox{\columnwidth}{!}{
        \tablestyle{2pt}{1.3}
        \begin{tabular}{x{35}y{82}|x{20}y{48}|x{72}}
            \multirow{2}{*}{Strategy} & \multicolumn{1}{c|}{\multirow{2}{*}{Configurations}} & \multicolumn{2}{c|}{\multirow{2}{*}{\shortstack{Heuristic Design\\(existing works)}}} & \multirow{2}{*}{\shortstack{\Ours\\(ours)}} \\
            & & & &  \\
            \shline
            \multirow{4}{*}{\shortstack{Generation\\Strategy}}& Re-masking Ratio $r(t)$ \color{ratio}  & \ \scriptsize  ${\color{ratio}r{(t)}}$ &\!\!\!\! \scriptsize $=\cos(\frac{\pi t}{2T})$ & \multirow{6}{*}{\shortstack{\shortstack{{{\shortstack{\hl{\scriptsize Jointly Optimized:}}}}\\$r^*(t),\tau_1^*(t),$\\$\tau_2^*(t),s^*(t),p^*(r)$}}}\\
            & Sampling Temp. \color{sampt}{$\tau_1(t)$} & \ \scriptsize   ${\color{sampt} \tau_1{(t)}} $ &\!\!\!\! \scriptsize $ = 1.0$ \\
            & Re-masking Temp. \color{remaskt} $\tau_2(t)$   & \ \scriptsize ${\color{remaskt} \tau_2{(t)}} $&\!\!\!\! \scriptsize $= \frac{\lambda(T-t+1)}{T}$ \\
            & Guidance Scale \color{cfg} $s(t)$ & \ \scriptsize  ${\color{cfg} s{(t)}}$&\!\!\!\! \scriptsize $=\frac{kt}{T}$ \\\cline{1-4}
            \multirow{2}{*}{\shortstack{Training\\Strategy}} & \multirow{2}{*}{Mask Ratio Dist. $p(r)$ } & \multirow{2}{*}{\ \scriptsize $p(r)$} & \multirow{2}{*}{\!\!\!\! \scriptsize $=\frac{2}{\pi\sqrt{1-r^2}}$} & \\
            & &&&\\\hline
            \multicolumn{5}{c}{\textbf{FID on ImageNet 256$\times$256} (model size $=$ 46M)} \\[-.5ex]
        \end{tabular}
    }
    \resizebox{\columnwidth}{!}{
        \tablestyle{2pt}{1.3}
        \begin{tabular}{x{121}|x{34}x{34}|x{34}x{34}}
            Generation Steps & $T=4$ & $T=8$ & $T=4$ & $T=8$ \\\hline
            TFLOPs / Image & 0.18 & 0.28 & 0.18 & 0.28 \\
            FID-50K & 8.40 & 5.73 & \dt{4.30}{$\downarrow$4.10} & \dt{3.52}{$\downarrow$2.21} \\
        \end{tabular}
    }
    \vskip -0.1in
    \caption{
        \textbf{Comparisons between existing works and \Ours}.
        \Ours tackles the strategy design of NATs by directly solving for the optimal solution, outperforming the heuristic-based counterparts by large margins.
        Here, $T$ denotes generation steps.
                The generation strategy is controlled by the scheduling functions $r(t)$, $\tau_1(t)$, $\tau_2(t)$, $s(t)$, while the training strategy is parameterized by a mask ratio distribution $p(r)$ (see Section ~\ref{sec:preliminaries} for details).
        \label{tab:teaser}
    }
%    \vskip -0.1in
\end{table}

%% file: sec/2_related.tex
\section{Related Work}

\paragraph{Image generation models} have historically been dominated by Generative Adversarial Networks (GANs)~\cite{goodfellow2014generative,brock2018large,Karras2019stylegan2}. Despite their success, GANs are challenging to train, prone to mode collapse, and heavily dependent on precise hyperparameter tuning and regularization~\cite{brock2018large,miyato2018spectral,brock2016neural,kang2023scaling}. Consequently, likelihood-based models such as diffusion~\cite{ramesh2021zero,ramesh2022hierarchical,dhariwal2021diffusion,saharia2022photorealistic,rombach2022high} and autoregressive models~\cite{yu2022scaling,ramesh2022hierarchical,yu2023scaling} have emerged. These models are gaining prominence due to their straightforward training processes, scalability, and capabilities to generate diverse, high-resolution images. However, their iterative refinement approach for sampling, while powerful, demands substantial computational resources, presenting significant hurdles for real-time applications and deployment on edge devices with limited computational capacity or  low latency is imperative.

\paragraph{Efficient image synthesis techniques} have seen numerous advancements recently.
There are primarily two approaches. The first involves developing new models that inherently support efficient sampling, such as non-autoregressive Transformers, which we discuss further in the next paragraph. The second approach focuses on enhancing existing models, particularly diffusion models, given their widespread use.
For example, DDIM~\cite{song2020denoising} and DPM-Solver~\cite{lu2022dpm,lu2022dpmp} have expedited diffusion sampling through intricate mathematical analysis, though they face challenges in preserving image quality with fewer than 10 sampling steps.
Other techniques, such as DDSS~\cite{watson2021learning} and AutoDiffusion~\cite{li2023autodiffusion}, have adopted an optimization-based approach to boost generation efficiency. Our research aligns with these efforts in its embrace of optimization, but it uniquely focuses on non-autoregressive Transformers for their inherent efficiency advantages. Furthermore, our approach examines the interplay between training and generation, enhancing both in a unified manner.
More recently, distillation-based methods~\cite{luo2023latent,sauer2023adversarial,yin2023one} have further reduced the sampling steps of diffusion models by transferring knowledge from a large pre-trained diffusion teacher to a few-step student generator.
Notably, this distillation approach to transferring the knowledge of a large diffusion teacher into a fewer-step student generator does not impose architectural constraints on the student generator.
Consequently, this technique is conceptually orthogonal to non-autoregressive Transformers and, by extension, remains compatible with the methodologies employed in \Ours.
\paragraph{Non-autoregressive Transformers (NATs)} originate from machine translation~\cite{ghazvininejad2019mask, gu2020fully}, known for their rapid inference capabilities.
Recently, these models have been applied to image synthesis, offering high-quality images with a minimal number of inference steps~\cite{chang2022maskgit,lezama2022improved,li2023mage,chang2023muse,qian2023strait}.
As a pioneering work, MaskGIT~\cite{chang2022maskgit} demonstrates highly-competitive fidelity and diversity on the ImageNet~\cite{russakovsky2015imagenet} with only 8 sampling steps.
It has been further extended for text-to-image generation and scaled up to 3B parameters in Muse~\cite{chang2023muse} and yields remarkable performance.
Separately, Token-critic~\cite{lezama2022improved} and MAGE~\cite{li2023mage} build upon NATs: Token-critic~\cite{lezama2022improved} enhances MaskGIT's performance by introducing an auxiliary model for guided sampling, while MAGE~\cite{li2023mage} proposes leveraging NATs to unify representation learning with image synthesis.
More recently, MAGVIT-v2~\cite{yu2023language} further advances the field with an enhanced tokenizer design.
Given that \Ours focuses on refining the training and generation strategies, which are the core components of all NATs, it inherently maintains compatibility with these approaches.

%% file: sec/3_1_preliminaries.tex
\section{On the Limitations of Non-Autoregressive Transformers (NATs)}
\subsection{Preliminaries of NATs}
\label{sec:preliminaries}
In this subsection, we start by giving an overview of the non-autoregressive Transformers~\cite{chang2022maskgit,chang2023muse,li2023mage} in image generation, laying the basis for our proposed method.
Non-autoregressive Transformers typically work in conjunction with a pre-trained VQ-autoencoder~\cite{van2017neural,razavi2019generating,esser2021taming} to generate images.
The VQ-autoencoder is responsible for the conversion between images and latent visual tokens, while the non-autoregressive Transformer learns to generate visual tokens in the latent VQ space.
As the VQ autoencoder is usually pre-trained and remains unchanged throughout the generation process, we mainly outline the training and generation strategies of non-autoregressive Transformers.

\paragraph{Training strategy.}
The training of non-autoregressive Transformers is based on the masked token modeling objective~\cite{devlin2019bert,bao2021beit,he2022masked}.
Specifically, we denote the visual tokens obtained by the VQ-encoder as $\bm{V}=[V_i]_{i=1:N}$, where $N$ is the sequence length.
Each visual token $v_i$ corresponds to a specific index of the VQ-encoder's codebook.
During training, a variable number of tokens, specifically \( \lceil r \cdot N \rceil \), are randomly selected and replaced with \texttt{[MASK]} token, where \( r \) is a mask ratio sampled from a predefined distribution \( p(r) \) within the $[0, 1]$ range.
The training objective is to predict the original tokens based on the surrounding unmasked ones, optimizing a cross-entropy loss function.

\paragraph{Generation strategy.}
During inference, non-autoregressive Transformers generate the latent visual tokens in a multi-step manner.
The model starts from an all-\texttt{[MASK]} token sequence $\bm{V}^{(0)}$.
At $t^{\textnormal{th}}$ step, the model predicts $\bm{V}^{(t)}$ from $\bm{V}^{(t-1)}$ by first \textbf{parallely decoding} all tokens and then \textbf{re-masking} less reliable predictions, as described below.

\begin{enumerate}
    \item \textbf{Parallel decoding.}  Given visual tokens $\bm{V}^{(t-1)}$, the model first parallely decodes all of the \texttt{[MASK]} tokens to form an initial guess $\hat{\bm{V}}^{(t)}$:
    \begin{equation*}
        \hat{V}^{(t)}_i
        \begin{cases}
            \sim  \hat {p}_{{\color{sampt}\tau_1{(t)}}}(V_i|\bm{V}^{{(t-1)}}),  & \text{if $V_{i}^{(t-1)} = $\texttt{[MASK]}}; \\
            = V^{(t-1)}_i,  & \text{otherwise}.
        \end{cases}
    \end{equation*}
    Here, ${\color{sampt}\tau_1{(\cdot)}}$ is the sampling temperature scheduling function, and $\hat {p}_{{\color{sampt}\tau_1{(t)}}}(V_i|\bm{V}^{{(t-1)}})$ represents the model's predicted probability distribution at position $i$, scaled by a temperature ${\color{sampt}\tau_1{(t)}}$.
    Meanwhile, confidence scores $\bm{C}^{(t)}$ are defined for all tokens:
    \begin{equation*}
        {C}_i^{(t)} =
        \begin{cases}
            \log \hat{p}(V_i=\hat{V}_i^{(t)}|\bm{V}^{(t-1)}),  & \text{if $V_{i}^{(t-1)} = $\texttt{[MASK]}};  \\
            +\infty,  & \text{otherwise}.
        \end{cases}
    \end{equation*}
    where $\hat{p}(V_i=\hat{V}_i^{(t)}|\bm{V}^{(t-1)})$ is the predicted probability for the selected token $\hat{V}_i^{(t)}$ at position $i$.
    \item \textbf{Re-masking.}  From the initial guess $\hat{\bm{V}}^{(t)}$, the model then obtains $\bm{V}^{(t)}$ by re-masking the $\lceil {\color{ratio}r{(t)}}\cdot N \rceil$ least confident predictions:
    \begin{equation*}
        V^{(t)}_i =
        \begin{cases}
            \hat{V}^{(t)}_i, & \text{if $i \in \mathcal{I}$}; \\
            \texttt{[MASK]} ,  & \text{if $i \notin \mathcal{I}$}.
        \end{cases}
    \end{equation*}
    Here, \( {\color{ratio}r{(\cdot)}} \in [0, 1] \) is the re-masking scheduling function, which regulates the proportion of tokens to be re-masked at each step. The set \(\mathcal{I}\) comprises indices of the \(N-\lceil {\color{ratio}r{(t)}}\cdot N \rceil\) most confident predictions and are sampled without replacement from $\text{Softmax}(\bm{C}^{(t)}/\tau_2(t))$\footnote{In practice, this sampling procedure is implemented via Gumbel-Top-$k$ trick~\cite{kool2019stochastic}.}, where $\color{remaskt}\tau_2{(\cdot)}$ is the re-masking temperature scheduling function.
\end{enumerate}
The model iterates the process for \( T \) steps to decode all \texttt{[MASK]} tokens, yielding the final sequence \( \bm{V}^{(T)} \).
The sequence is then fed into the VQ-decoder to obtain the image.

\input{sec/3_2_limitations.tex}

%% file: sec/3_2_limitations.tex
\subsection{Limitations of NATs}
\label{sec:limitations}

Compared to other likelihood-based generative models, one of the predominant advantages of NATs is their superior efficiency~\cite{chang2022maskgit,chang2023muse}. Once they are appropriately deployed, decent-quality images can be generated with a few sampling steps. However, it is usually non-trivial for practitioners to utilize these models properly, since their performance tends to heavily depend on multiple scheduling functions that need to be carefully configured.
As discussed above, a generation process typically involves three scheduling functions to control the mask ratio $r(t)$, the sampling temperature $\tau_1(t)$, and the re-masking temperature $\tau_2{(t)}$, respectively.
When classifier-free guidance~\cite{dhariwal2021diffusion,chang2023muse} is adopted, a guidance scale scheduling function $\color{cfg}s{(t)}$ is further introduced~\cite{chang2023muse} to progressively adjust the guidance strength.
Existing works typically design these functions with heuristic rules (see Table~\ref{tab:teaser}), necessitating expert knowledge and manual effort.
Meanwhile, these heuristic rules may fall short of capturing the optimal dynamics of the generation process, leading to sub-optimal designs (see Table~\ref{tab:contribution}).

\begin{figure}[t!]\centering
\includegraphics[width=\linewidth]{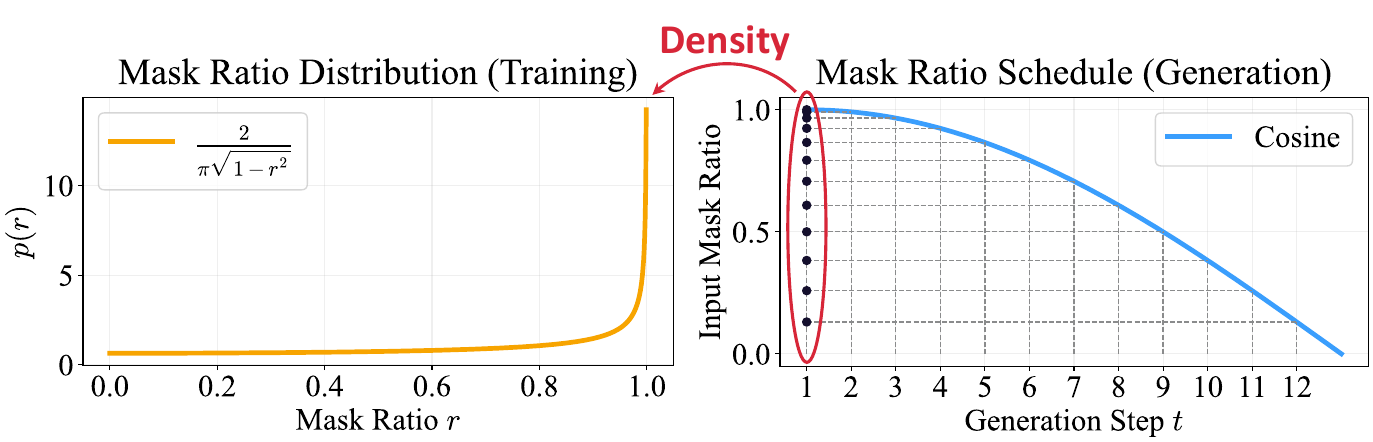}
\vskip -0.1in
\caption{
    \textbf{The heuristic design of $p(r)$ in existing works}:
    the density of $p(r)$ reflects the frequency of mask ratios encountered during generation.
    We take $T\!=\!12$ for example. Notably, as shown in Table~\ref{tab:comp_fixed}, such a heuristic design is sub-optimal.
}
\vskip -0.1in
\label{fig:heuristic_for_train_dist}
\end{figure}
\begin{table}[t]
    \centering
    \tablestyle{2.4mm}{1.5}
    \begin{tabular}{c|cccc|c}
        Training Distribution $p(r)$ &  Original: \raisebox{1ex}{$\frac{2}{\pi\sqrt{1-r^2}}$}   & Fixed: $r\equiv0.8$ \\\shline
        FID-50K &  8.40 & \textbf{8.31}  \\
    \end{tabular}
    \vskip -0.1in
    \caption{
        \textbf{Heuristically-designed $p(r)$ in existing works \vs fixed mask ratio} for training NATs.
        A simple fixed mask ratio produces even better results.\label{tab:comp_fixed}
    }
    \vskip -0.2in
\end{table}

Moreover, this intricate generation process requires the support of a well-designed training strategy.
More precisely, it is essential to design a proper $p(r)$ for mask ratio sampling during training, such that NATs can effectively process diverse input distributions with varying proportions of mask tokens during generation.
In most existing works~\cite{chang2022maskgit,chang2023muse,qian2023strait}, $p(r)$ is configured to mimic the variation of mask ratios in the generation process, as illustrated in Figure~\ref{fig:heuristic_for_train_dist}.
However, this common design may not be proper.
A counterintuitive finding in Table~\ref{tab:comp_fixed} reveals that even using a \emph{single, fixed} mask ratio can produce \emph{better} results than the heuristic design.
A possible explanation is that the capabilities model learned from decoding at one mask ratio are highly transferable for decoding other ratios of mask tokens.
Consequently, the strict, proportional allocation of \( p(r) \)'s density, based on the encountered frequencies of mask ratios during generation, could be sub-optimal.
This finding suggests the need for a more systematic approach to better capture the relationship between training and inference.

%% file: sec/3_3_method.tex
\section{Method}
\label{sec:method}
As aforementioned, the heuristic-driven design of training and generation strategies for non-autoregressive generative models is both labor-intensive and sub-optimal.
To address this issue, we propose an optimization-based approach to derive these optimal configurations with minimal human effort.
In this section, we elaborate on our \Ours method.

\subsection{A Unified Optimization Framework}
\paragraph{Mathematical formulation.}
Our objective is to determine the most appropriate configurations for the training and generation procedures of non-autoregressive Transformers.
From the lens of optimization, this corresponds to identifying the optimal generation scheduling functions \( r^*(t) \), \( \tau_1^*(t) \), \( \tau_2^*(t) \), \( s^*(t) \) and mask ratio distribution \( p^*(r) \) that maximize the performance of the model according to a chosen metric.
Notably, since a generation process of $T$ steps involves $T$ values of each scheduling function, we can circumvent the difficulty of directly optimizing the functions by optimizing four groups of hyperparameters \( \bm{r}= [r{(t)}]_{t=1:T} \), \( \bm{\tau_1} = [\tau_1{(t)}]_{t=1:T} \), \( \bm{\tau_2} = [\tau_2{(t)}]_{t=1:T} \), \( \bm{s} = [s{(t)}]_{t=1:T} \) instead.
Formally, this optimization problem can be defined as:
\setlength{\abovedisplayskip}{-9pt}
\setlength{\belowdisplayskip}{-1pt}

\begin{align}
    &\bm{r^*}, \bm{\tau_1^*}, \bm{\tau_2^*}, \bm{s^*}, p^*(r) = \argmax_{\bm{r}, \bm{\tau_1}, \bm{\tau_2}, \bm{s}, p(r)} F\left(\bm{r}, \bm{\tau_1}, \bm{\tau_2}, \bm{s}, \bm{\theta}^*_{p(r)}\right), \nonumber \\
    &\text{s.t.}\phantom{p^*(r),} \bm{r} \in [0, 1]^T, \bm{\tau_1}, \bm{\tau_2}, \bm{s} \in \mathbb{R}^T_+, \nonumber \\
    &\phantom{\text{s.t.} p^*(r),} p(r) \ge 0 \quad \forall r \in [0, 1], \int_0^1 p(r) dr = 1. \nonumber
\end{align}

\setlength{\abovedisplayskip}{5pt}
\setlength{\belowdisplayskip}{5pt}
Herein, \( \bm{\theta}_{p(r)}^* \) denotes the model parameters trained under the mask ratio distribution \( p(r) \).
The function \( F \) measures the generation quality, and can be instantiated with metrics like Fréchet inception distance (FID)~\cite{heusel2017gans}, Inception score (IS)~\cite{salimans2016improved}, \emph{etc}.
Note that for metrics like FID, which are considered better when smaller, we maximize their negative values.
In addition, the constraints on \( p(r) \) confirm that it's a valid probability distribution.

\paragraph{Alternating optimization.}
Before solving the optimization problem, we note an important distinction between \( p(r) \) and other variables: \( p(r) \) is nested within model parameter term \( \bm{\theta}^*_{p(r)} \).
Consider an arbitrary optimization procedure, where we obtain an intermediate candidate for \( p(r) \). It is usually essential to train the model with \( p(r) \) and evaluate how good \( p(r) \) is. In contrast, evaluating $\bm{r}, \bm{\tau_1}, \bm{\tau_2}, \bm{s}$ only necessitates inferring the generative model, which is much cheaper than \( p(r) \).
As a consequence, directly solving all variables simultaneously would be inefficient as the slow evaluation of \( p(r) \) hinders the optimization of other variables, which could otherwise proceed at a much faster pace.

Motivated by the imbalanced nature of the variables, we divide our optimization problem into two sub-problems: generation strategy optimization and training strategy optimization, and propose to solve them with an alternating algorithm. The first sub-problem focuses on obtaining the optimal configuration for the hyperparameters controlling the generation procedure \( \bm{r}, \bm{\tau_1}, \bm{\tau_2}, \bm{s} \), given a model trained under \( p(r) \), while the second sub-problem optimizes \( p(r) \) with the given generation configurations. These two sub-problems are solved alternatively until the generation quality of the model converges. This iterative approach allows the rapid adjustment of \( \bm{r}, \bm{\tau_1}, \bm{\tau_2}, \bm{s} \), while steadily steering \( p(r) \) towards optimality, yielding a superior efficiency for solving the overall optimization problem. The empirical evidence can be found in Table~\ref{tab:alter_concurrent}.

%\vspace{-3ex}
\begin{center}
    \resizebox{.9\columnwidth}{!}{
        \begin{minipage}{\linewidth}
        \begin{algorithm}[H]
            \caption{Alternating Algorithm for Solving the Training \& Inference Configurations of NATs\label{alg:uni_optimization}}
            \begin{algorithmic}[1]
                \State Initialize \( p(r), \bm{r}, \bm{\tau_1}, \bm{\tau_2}, \bm{s} \)
                \State Set convergence threshold \( \epsilon \), \( F_{\text{prev}} \gets \infty \)
                \Repeat
                    \State \textcolor{blue}{\texttt{\# Alternating optimization}}
                    \State \( \bm{r}, \bm{\tau_1}, \bm{\tau_2}, \bm{s} \gets \text{OptimizeGenerationStrategy}(p(r)) \)
                    \State \( p(r) \gets \text{OptimizeTrainingStrategy}(\bm{r}, \bm{\tau_1}, \bm{\tau_2}, \bm{s}) \)
                    \State \textcolor{blue}{\texttt{\# Evaluate the strategies}}
                    \State \( \bm{\theta}^*_{p(r)} \gets \text{TrainModel}(p(r)) \)
                    \State \( F_{\text{new}} \gets \text{Evaluate}(\bm{r}, \bm{\tau_1}, \bm{\tau_2}, \bm{s}, \bm{\theta}^*_{p(r)}) \)
                    \State \( \Delta F \gets |F_{\text{new}} - F_{\text{prev}}| \)
                    \State \( F_{\text{prev}} \gets F_{\text{new}} \)
                \Until{\( \Delta F \leq \epsilon \)}
                \State \Return \( \bm{r}, \bm{\tau_1}, \bm{\tau_2}, \bm{s}, p(r) \)
            \end{algorithmic}
        \end{algorithm}
        \end{minipage}
    }
\end{center}
We summarize our overall optimization process in Algorithm~\ref{alg:uni_optimization}.
In the following, we elaborate on the optimization of the generation strategy and training strategy respectively.

\subsection{Generation Strategy Optimization}
In this sub-problem, we are interested in searching for the optimal generation strategy variables \( \bm{r}, \bm{\tau_1}, \bm{\tau_2}, \bm{s} \) given a fixed \( p(r) \), which can be formulated as:
\setlength{\abovedisplayskip}{5pt}
\setlength{\belowdisplayskip}{5pt}
\begin{align*}
    &\bm{r^*}, \bm{\tau_1^*}, \bm{\tau_2^*}, \bm{s^*} = \argmax_{\bm{r}, \bm{\tau_1}, \bm{\tau_2}, \bm{s}} F(\bm{r}, \bm{\tau_1}, \bm{\tau_2}, \bm{s}, \bm{\theta}_{p(r)}^*), \\
    &\text{s.t.} \quad \bm{r} \in [0, 1]^T, \bm{\tau_1}, \bm{\tau_2}, \bm{s} \in \mathbb{R}^T_+.
\end{align*}
Inspired by the success of gradient-based optimization in deep learning~\cite{ruder2016overview}, we find this sub-problem can be effectively solved via simple gradient descent.
In specific, although $F$ is not differentiable with respect to \( \bm{r}, \bm{\tau_1}, \bm{\tau_2}, \bm{s} \) due to the parallel decoding process (see Section~\ref{sec:preliminaries}),
it is feasible to approximate the gradients corresponding to each variable by leveraging a finite difference method.
\input{tables/main_results/imnet.tex}
Let \( \bm{\xi} = [\bm{r}, \bm{\tau_1}, \bm{\tau_2}, \bm{s}] \) be the concatenation of variables. The gradient of \( F \) with respect to \( \bm{\xi} \) can be approximated by
\[
    \frac{\partial F(\bm{\xi}, \bm{\theta}^*_{p(r)})}{\partial \xi_i} \approx \frac{F(\bm{\xi} + \epsilon \bm{e}_i, \bm{\theta}^*_{p(r)}) - F(\bm{\xi}, \bm{\theta}^*_{p(r)})}{\epsilon},
\]
where \( i \) ranges over the dimensions of \( \bm{\xi} \), \( \bm{e}_i \) is the unit vector in the direction of \( \xi_i \), and \( \epsilon \) is a small positive number. We denote the estimated gradients as \( \hat{\nabla}_{\bm{\xi}} F(\bm{\xi}, \bm{\theta}^*_{p(r)}) \). Then the hyperparameters can be updated by gradient descent:
\[
\bm{\xi} \gets \bm{\xi} - \eta \hat{\nabla}_{\bm{\xi}} F(\bm{\xi}, \bm{\theta}^*_{p(r)}),
\]
where \( \eta \) denotes the learning rate.

\subsection{Training Strategy Optimization}
In this sub-problem, we focus on searching for the optimal mask ratio distribution \( p^*(r) \) with a given generation strategy \( \bm{r}, \bm{\tau_1}, \bm{\tau_2}, \bm{s} \).
Notably, since we usually need to train the model to evaluate any candidate values of \( p^*(r) \), optimizing $p(r)$ in a general probability distribution space is computationally expensive.
As a result, we propose to restrict $p(r)$ to a specific family of probability density functions.
In this work, we adopt the Beta distribution as the family of $p(r)$:
\[
    p(r; \alpha, \beta) = \frac{r^{\alpha - 1} (1 - r)^{\beta - 1}}{B(\alpha, \beta)},
\]
where $B(\alpha, \beta)$ is the Beta function and $\alpha, \beta > 0$.
Thus, we can simplify our optimization problem as:
\begin{align*}
    &\alpha^*, \beta^* = \argmax_{\alpha, \beta} F(\bm{r}, \bm{\tau_1}, \bm{\tau_2}, \bm{s}, \bm{\theta}_{p(r; \alpha, \beta)}^*), \\
    &\text{s.t.} \quad \alpha, \beta > 0.
\end{align*}
With such an assumption, we find it is feasible to effectively solve \( p^*(r) \) with a simple greedy search algorithm~\cite{boyd2004convex}.
Specifically, we begin by performing a line search to optimize one parameter while keeping the other fixed. Once we find an optimal value for the first parameter, we switch to the second parameter and conduct a line search again to find its optimal value. This optimization continues until there is no further improvement in the performance metric.

%% file: tables/main_results/imnet.tex
\begin{table*}[!t]
    \centering
    \tablestyle{5pt}{1.1}
    \resizebox{2.08\columnwidth}{!}{
        \begin{tabular}{y{100}x{22}x{26}x{17}x{25}x{25}x{20}|x{26}x{17}x{25}x{25}x{20}}
            \multirow{2}{*}{Method} & \multirow{2}{*}{Type} & \multicolumn{5}{c}{\textbf{ImageNet-256}} & \multicolumn{5}{c}{\textbf{ImageNet-512}} \\
            & & \#Params & Steps & TFLOPs$\downarrow$ & FID$\downarrow$ & IS$\uparrow$  & \#Params & Steps & TFLOPs$\downarrow$ & FID$\downarrow$ & IS$\uparrow$ \\\shline
            \VQVAE             & AR    & 13.5B & 5120    & -    & 31.1  & $\sim$ 45 & -    & -       & -     & -     & -      \\
            \VQGAN             & AR    & 1.4B  & 256     & -    & 15.78 & 78.3      & 227M & 1024    & -     & 26.52 & 66.8   \\
            ADM-G~\cite{dhariwal2021diffusion}\pub{NeurIPS'21}               & Diff. & 554M  & 250     & 334.0  & 4.59  & 186.7     & 559M & 250     & 579.0 & 7.72  & 172.7 \\
            ADM-G, ADM-U~\cite{dhariwal2021diffusion}\pub{NeurIPS'21} & Diff. & 608M  & 250 & 239.5    & 3.94  & 215.8    & 731M & 250 & 719.0 & 3.85  & 221.7  \\
            \LDM               & Diff. & 400M  & 250     & 52.3 & 3.60  & 247.7    & -    & -       & -     & -     & -      \\
            \VQDiffusion       & Diff. & 554M  & 100     & 12.4 & 11.89  & -         & -    & -       & -     & -     & -      \\
            \DraftRevise       & NAT   & 1.4B  & 72      & -    & 3.41  & 224.6     & -    & -       & -     & -     & -      \\\hline\multicolumn{10}{l}{\emph{Efficient likelihood-based image generation models \emph{($^\dagger$: augmented with \Solver)}}} \\\hline
            \multirow{2}{*}{\ADMp} & \multirow{2}{*}{Diff.} & \multirow{2}{*}{554M} & 4 & 5.3 & 22.35 & - & \multirow{2}{*}{559M} & 4 & 9.3 & 42.85 & - \\
            & & & 16 & 21.4 & 5.28 & 214.8 &  & 16 & 37.1 & 8.8 & 157.2 \\
            \multirow{2}{*}{\LDMp} & \multirow{2}{*}{Diff.} & \multirow{2}{*}{400M} & 4 & 1.2 & 11.74 & - & \multirow{2}{*}{-} & - & - & - & - \\
            & & & 16 & 3.7 & 3.68 & 202.7 &  & - & - & - & -  \\
            \multirow{2}{*}{\UVITp} & \multirow{2}{*}{Diff.} & \multirow{2}{*}{501M} & 4 & 1.4 & 8.45 & - & \multirow{2}{*}{501M} & 4 & 2.3 & 8.29 & - \\
            & & & 16 & 4.6 & 2.77 & 259.5 &  & 16 & 5.5 & 4.04 & 252.6 \\
            \multirow{2}{*}{\DITp} & \multirow{2}{*}{Diff.} & \multirow{2}{*}{675M} & 4 & 1.3 & 9.71 & - & \multirow{2}{*}{675M} & 4 & 5.4 & 11.72 & - \\
            & & & 16 & 4.1 & 3.13 & 256.1 &  & 16 & 18.0 & 3.84 & 211.7 \\
            \MaskGIT & NAT & 227M & 8 & 0.6 & 6.18 & 182.1 & 227M & 12 & 3.3 & 7.32 & 156.0 \\
            \TokenCritic & NAT & 422M & 36 & 1.9 & 4.69 & 174.5 & 422M & 36 & 7.6 & 6.80 & 182.1  \\
            \gc{MaskGIT~\cite{chang2022maskgit}\pub{CVPR'22}} & \gc{NAT} & \gc{227M} & \gc{8} & \gc{-} & \gc{4.02} & \gc{-} & \gc{227M} & \gc{12} & \gc{-} & \gc{4.46} & \gc{-} \\
            \gc{Token-Critic~\cite{lezama2022improved}\pub{ECCV'22}} & \gc{NAT} & \gc{422M} & \gc{36} & \gc{-} & \gc{3.75} & \gc{-} & \gc{422M} & \gc{36} & \gc{-} & \gc{4.03} & \gc{-}  \\
            \MAGE & NAT & 230M & 20 & 1.0 & 6.93 & - & -    & -       & -     & -     & -    \\\hline
            \Ours-L$^\ddagger$ & NAT & 194M & 12 & 0.7 & 4.45 & 193.3 & 199M & 12 & 1.0  & 6.36 & 185.0 \\
            \multirow{2}{*}{\Ours-S} & \multirow{2}{*}{NAT} & \multirow{2}{*}{46M} & 4 & \textbf{0.2} & 4.30 & 249.7  &\multirow{2}{*}{48M}  & 4 & \textbf{0.5} & 6.59 & 252.9 \\
            & & & 8 & 0.3 & 3.52 & 253.4  &  & 8 & 0.6 & 5.06 & 254.5 \\
            \multirow{2}{*}{\Ours-L} & \multirow{2}{*}{NAT} & \multirow{2}{*}{194M} & 4 & 0.5 & 3.26 & 271.3  & \multirow{2}{*}{199M} & 4 & 0.8  &  5.37 & 278.7 \\
            & & & 8 & 0.9 & \textbf{2.68} &   278.8 &  & 8 & 1.2 & \textbf{3.74} & 286.2 \\
        \end{tabular}
    }
    \vskip -0.1in
    \caption{
        \textbf{Class-conditional image generation on ImageNet-256 and ImageNet-512\label{tab:fid_imnet}}.
        TFLOPs represents the number of floating-point operations required for generating a single image.
        We calculate FID-50K following~\cite{dhariwal2021diffusion,peebles2023scalable}.
        For DPM-Solver~\cite{lu2022dpm} augmented diffusion models (marked with $^\dagger$), we follow instructions in~\cite{lu2022dpm} to tune all solver configurations as well as classifier-free guidance scale, and report results with the lowest FID.
        $^\ddagger$: methods without classifier-based or classifier-free guidance~\cite{dhariwal2021diffusion,ho2022classifier}.
        \gc{Rows in gray} use compute-intensive classifier-based rejection sampling.
        Diff: diffusion, AR: autoregressive, NAT: non-autoregressive Transformers.
    }
    \vskip -0.1in
\end{table*}

%% file: sec/4_experiment.tex
\section{Experiments}
\label{sec:experiments}

\paragraph{Implementation details.}
Following~\cite{chang2022maskgit,li2023mage,chang2023muse}, we employ a pretrained VQGAN~\cite{esser2021taming} with a codebook of size 1024 for the conversion between images and visual tokens.
The architecture of our models follows U-ViT~\cite{bao2022all}, a type of Transformer adapted for image generation tasks.
We consider two model configurations: a small model (13 layers, 512 embedding dimensions, denoted as \Ours-S) and a large model (25 layers, 768 embedding dimensions, denoted as \Ours-L). For ImageNet-512, we adopt a patch size of 2 to accommodate the increased token count.
In the implementation of \Ours, we first adopt the small model and perform alternating optimization on ImageNet-256 ($T=4$) for three iterations to obtain our basic training and generation strategy, where we utilize FID~\cite{heusel2017gans} as the default evaluation metric $F$.
When applying the basic strategy to different datasets and network architectures, we conduct a single additional search for only the generation strategy. This implementation technique slightly improves the performance of \Ours by fine-tuning it conditioned on the specific scenario (see Table \ref{tab:transferability} for ablation studies). Notably, adjusting the generation strategy is more efficient since it only needs to infer the model.

\subsection{Main Results}
\label{sec:main_res}

\paragraph{Class-conditional generation on ImageNet.}
In Table~\ref{tab:fid_imnet}, we present a comparison of our method against state-of-the-art generative models on ImageNet 256x256 and ImageNet 512x512.
The number of generation steps, model parameters and the total computational cost during generation (measured in TFLOPs) are also reported to give a comprehensive evaluation of the efficiency-effectiveness tradeoff.
Moreover, in Table~\ref{tab:fid_imnet}, the models specialized in efficient image synthesis, \emph{e.g.}, recently proposed non-autoregressive Transformers and diffusion models with advanced samplers~\cite{lu2022dpm,lu2022dpmp}, are grouped together for a direct comparison with our method.
\begin{figure*}[t!]\centering
\includegraphics[width=.93\linewidth]{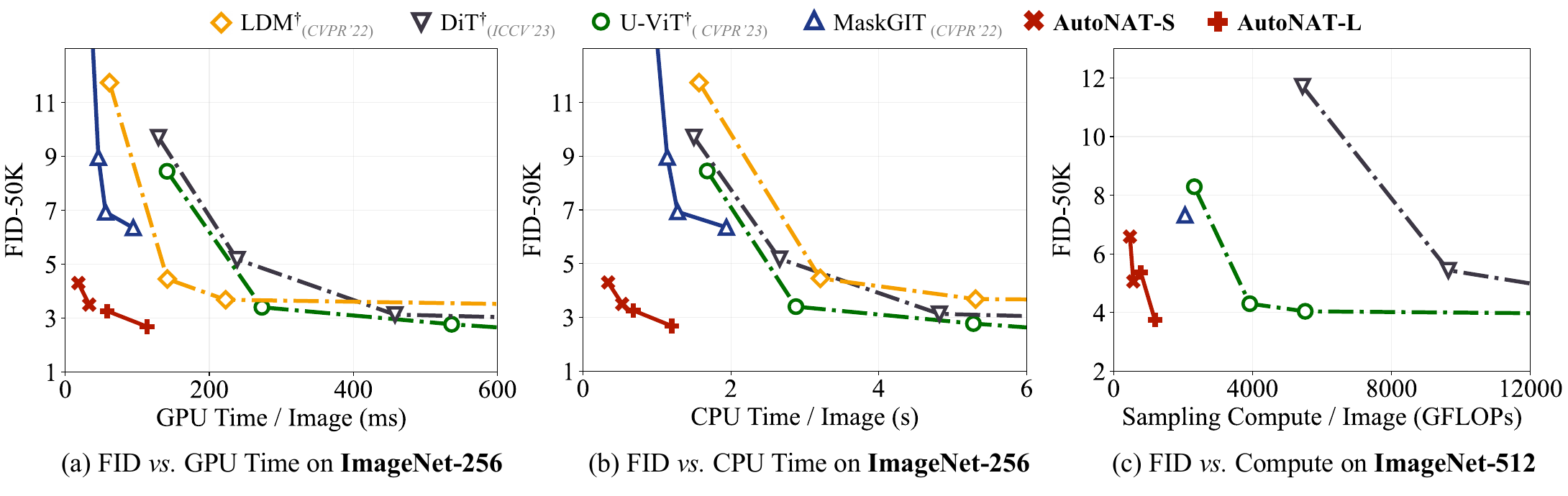}
\vskip -0.15in
\caption{
    \textbf{Sampling efficiency} on ImageNet-256 and ImageNet-512.\label{fig:tradeoffs_imnet}
    LDM is not included in ImageNet-512 results as it is only trained on ImageNet-256.
    GPU time is measured on an A100 GPU with batch size 50.
    CPU time is measured on Xeon 8358 CPU with batch size 1.
    $^\dagger$: DPM-Solver~\cite{lu2022dpm} augmented diffusion models.
}
\label{fig:tradeoff_curves}
\vskip -0.12in
\end{figure*}

The results in Table~\ref{tab:fid_imnet} show that \Ours-S, although having notably fewer parameters than other baselines, yields competitive performance on ImageNet-256. For example, \Ours-S requires only 0.2 TFLOPs and 4 synthesis steps to achieve an FID of 4.30. With a slightly increased computational budget of 0.3 TFLOPs, the FID obtained by \Ours-S is further improved to 3.52, surpassing most of the baselines. Furthermore, the larger \Ours continues this trend, attaining an FID of 2.68 with 8 steps. On ImageNet-512, our best model achieves an FID of 3.74, exceeding other state-of-the-art models while utilizing significantly fewer computational resources.

\begin{figure*}[!t]\centering
\includegraphics[width=.9\linewidth]{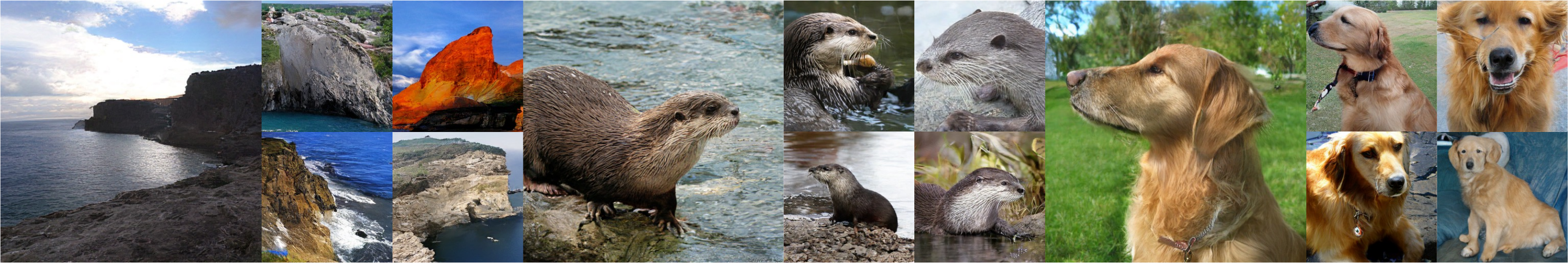}
\vskip -0.08in
\caption{
    \textbf{Selected visualizations of \Ours}. Samples are generated in 8 steps with \Ours-L on ImageNet-512 and ImageNet-256.
}
\label{fig:vis}
\vskip 0.05in
\end{figure*}

In addition, we present more comprehensive comparisons of the trade-off between generation quality and computational cost in Figs.~\ref{fig:tflops_imnet} and \ref{fig:tradeoff_curves}.
Importantly, both the theoretical GFLOPs and the practical GPU/CPU latency required for generating an image are reported.
The results demonstrate that our method consistently outperforms other baselines in terms of both generation quality and computational cost.
Compared to the strongest diffusion models with fast samplers, \Ours offers approximately a $5\times$ inference speedup without compromising performance. Qualitative results of our method are presented in Figure~\ref{fig:vis}.

\input{tables/main_results/mscoco.tex}
\input{tables/main_results/cc3m.tex}
\paragraph{Text-to-image generation on MS-COCO.}
We further evaluate the effectiveness of \Ours in the text-to-image generation scenario, using the MS-COCO~\cite{lin2014microsoft} benchmark. As summarized in Table~\ref{tab:fid_ms}, \Ours-S is able to outperform other baselines with a minimal 0.3 TFLOPs compute and achieves a leading FID score of 5.36, indicating its superior efficiency and effectiveness.
When compared to the latest diffusion models equipped with a fast sampler~\cite{bao2022all}, \Ours-S surpasses its 50-step sampling results while requiring 7$\times$ less computational cost, and outperforming it by large margins with similar computational resources.

\paragraph{Text-to-image generation on CC3M.}
In Table~\ref{tab:cc3m}, we validate the effectiveness of \Ours for text-to-image generation on the larger CC3M dataset~\cite{sharma2018conceptual}.
Here its efficiency is mainly evaluated on top of the recently proposed non-autoregressive Transformer model: Muse~\cite{chang2023muse}.
We apply \Ours to the pre-trained Muse model, and only optimize the generation strategy.
As indicated in Table~\ref{tab:cc3m}, combining our optimized strategy with the Muse model yields an FID of 6.90 on CC3M, surpassing other baselines notably.
Compared to the vanilla Muse model, \Ours outperforms it with half of the computational cost and achieves significantly better FID (6.90 \vs 7.67) when utilizing the same computational resources.

\subsection{Analytical Results}
In this section, we provide more analytical results of our method.
Unless otherwise specified, we adopt \Ours-S trained on ImageNet-256 in all experiments.
\label{sec:experiments_abl}

\input{tables/ablations}

\paragraph{Contribution of training and generation strategies.}
Our method optimizes both training and generation strategies by default, and we conduct a more fine-grained ablation study in Table \ref{tab:contribution} to assess the contribution of each component. Removing the optimized training strategy has a noticeable detrimental impact on the FID score (an increase of 0.91). However, ablating the optimized generation strategy has an even more significant negative effect, deteriorating the FID by a substantial 3.58. This indicates that while both components are important, the previous non-autoregressive models had more room for improvement in generation strategies

\paragraph{Optimization efficacy.}
Our \Ours algorithm is iterative, yet as evidenced in Table~\ref{tab:optim_process}, it rapidly converges to a decent solution within a few iterations. Compared to the baseline FID of 8.40, a single optimization iteration markedly improves the FID to 4.61. Within just three iterations, the algorithm largely converges, evidenced by a minimal FID difference of 0.05 between the last two iterations, culminating in a performance of 4.30 FID, substantially exceeding the baseline. This underscores \Ours's effectiveness in optimizing training and generation configurations for non-autoregressive Transformers.

\paragraph{Alternating \vs concurrent optimization.}
We compare our alternating optimization strategy against optimizing the training and generation strategy concurrently (denoted as Concurrent) in Table~\ref{tab:alter_concurrent}.
Optimization efficiency is quantified in terms of computational cost measured by training time on a single A100 node.
Our alternating optimization method demonstrates significantly better performance (4.30 \vs 8.01) under similar computational costs.
Even when the resource allocation for concurrent optimization was doubled, its performance still remained considerably worse than that of our method.
This observation demonstrates the effectiveness of our proposed alternating optimization in facilitating more rapid and robust convergence.

\paragraph{Transferability of the searched strategies.}
In Table~\ref{tab:transferability}, we examine the transferability of strategies developed on our smaller model to larger models. To investigate this issue comprehensively, we introduced an additional ``base'' model (consisting of 12 layers with an embedding dimension of 768) and trained it with the same configuration as other models on ImageNet-256. The empirical results indicate that strategies formulated for smaller models are not only transferable but also highly effective when applied to larger models. Both directly transferred and re-optimized strategies for larger models significantly outperform the baseline at all generation steps. The re-optimized strategies show a slight advantage (about 0.1) over directly transferred strategies. This minor difference underscores the robust generalization capacity of the strategies across varying model sizes.

\paragraph{Impacts of each generation hyperparameter.}
Finally, we evaluated the impact of each hyperparameter by omitting one from our optimization set. As demonstrated in Table~\ref{tab:abl_upd_set}, excluding any hyperparameter variably affects the model's performance. The re-masking ratios \(\bm{r}\) exhibit a minor influence on performance, with a 0.28-point increase in FID when left unoptimized. Conversely, the most significant impact is observed when the sampling temperature \(\bm{\tau_1}\) remains unoptimized, leading to a 2.27-point increase in FID. These findings indicate the extent of sub-optimality in current heuristically designed generation hyperparameters and may offer valuable insights for developing enhanced generation strategies in non-autoregressive Transformers.

%% file: tables/main_results/mscoco.tex
\begin{table}[t!]
    \centering
    \resizebox{\linewidth}{!}{
        \tablestyle{4pt}{1.15}
        \begin{tabular}{y{70}x{22}x{26}x{17}x{25}x{30}}
            Method & Type & \#Params &  Steps & TFLOPs$\downarrow$ & FID$\downarrow$\\\shline
            VQ-Diffusion~\cite{gu2022vector} & Diff. & 370M & 100 & - & 13.86 \\
                Frido~\cite{fan2023frido} & Diff. &512M & 200 & - & 8.97 \\
            U-Net$^\dagger$~\cite{bao2022all} & Diff. &53M & 50 & - & 7.32 \\
                                        U-ViT$^\dagger$~\cite{bao2022all} & Diff. & 58M & 4 & 0.4 & 11.88 \\
            U-ViT$^\dagger$~\cite{bao2022all} & Diff. & 58M & 50 & 2.2 & 5.48 \\\hline
                        \Ours-S & NAT & 45M & 8 &  \textbf{0.3} & \textbf{5.36} \\
        \end{tabular}
    }
    \vskip -0.1in
    \caption{
        \textbf{Text-to-image generation on MS-COCO\label{tab:fid_ms}}; all models are trained and evaluated on MS-COCO.
        We report FID-30K following~\cite{bao2022all}.
        $^\dagger$: augmented with DPM-Solver~\cite{lu2022dpm,lu2022dpmp}.
    }
    \vskip -0.1in
\end{table}

%% file: tables/main_results/cc3m.tex
\begin{table}[t!]
    \centering
    \resizebox{\linewidth}{!}{
        \tablestyle{4pt}{1.15}
        \begin{tabular}{y{70}x{22}x{26}x{17}x{25}x{30}}
            Method                    & Type           & \#Params             & Steps & TFLOPs$\downarrow$ & FID$\downarrow$   \\\shline
            VQGAN~\cite{esser2021taming}                     & AR                   & 600M                  & 256                &   -    & 28.86 \\
            LDM-4~\cite{rombach2022high}                     & Diff.                & 645M                  & 50                  &   -    & 17.01 \\
            RQ-Transformer~\cite{lee2022autoregressive}            & AR                   & 654M                  & 64                  &   -    & 12.33 \\
            Draft-and-revise~\cite{lee2022draft}          & NAT                  & 654M                  & 72                  &   -    & 9.65  \\
                                    \multirow{2}{*}{Muse~\cite{chang2023muse}} & \multirow{2}{*}{NAT} & \multirow{2}{*}{500M} & 8                  &    2.8    & 7.67  \\
            &                      &                       & 16                 &    5.4    & 7.01  \\\hline
            \Ours-Muse~\cite{chang2023muse}             & NAT                  & 500M                  & 8                  &    \textbf{2.8}    & \textbf{6.90}  \\
        \end{tabular}
    }
    \vskip -0.1in
    \caption{
    \textbf{Text-to-image generation on CC3M\label{tab:cc3m}}; all models are trained and evaluated on CC3M.
    We report FID-30K following~\cite{chang2023muse}.
    }
    \vskip -0.2in
\end{table}

%% file: tables/ablations.tex
\begin{table*}[t]
    \vspace{-.2em}
    \centering
    \subfloat[
        \textbf{Optimization contribution}. Assessing the effect of optimized training and generation strategies on FID scores against baselines.
        \label{tab:contribution}
    ]{
        \centering
        \begin{minipage}{0.29\linewidth}{\begin{center}
                                             \tablestyle{1.2mm}{1.15}
                                             \begin{tabular}{ccl}
                                                 Train Strategy & Gen. Strategy & FID $\downarrow$ \\\shline\rowcolor{defaultcolor}
                                                 \cmark         & \cmark        & \textbf{4.30}    \\
                                                 & \cmark        & 5.21 \textcolor{gray}{\tiny{($+$0.91)}}            \\
                                                 \cmark         &               & 7.88 \textcolor{gray}{\tiny{($+$3.58)}}            \\
                                                 &               & 8.40 \textcolor{gray}{\tiny{($+$4.10)}}            \\
                                             \end{tabular}
        \end{center}}\end{minipage}
    }
    \hspace{2em}
    \subfloat[
        \textbf{Optimization efficacy}. Our optimization process quickly converges and surpasses the \gc{baseline results} by a large margin.
        \label{tab:optim_process}
    ]{
        \begin{minipage}{0.29\linewidth}{\begin{center}
                                             \tablestyle{1.2mm}{1.15}
                                             \begin{tabular}{cc}
                                                 Optimization Steps & FID $\downarrow$ \\\shline
                                                 \gc{-}             & \gc{8.40}        \\
                                                 1                  & 4.61             \\
                                                 2                  & 4.35             \\\rowcolor{defaultcolor}
                                                 3                  & \textbf{4.30}    \\
                                             \end{tabular}
        \end{center}}\end{minipage}
    }
    \hspace{2em}
    \subfloat[
        \textbf{Alternating \vs concurrent optimization}.
        Alternating optimization is faster and more effective.
        Cost measured in one A100 node.
        \label{tab:alter_concurrent}
    ]{
        \begin{minipage}{0.29\linewidth}{\begin{center}
                                             \tablestyle{8pt}{1.15}
                                             \begin{tabular}{ccc}
                                                 Method      & Cost $\downarrow$ & FID $\downarrow$ \\\shline\rowcolor{defaultcolor}
                                                 Alternating & 1.9 days          & \textbf{4.30}    \\
                                                 Concurrent       & 2.0 days          & 8.01             \\
                                                 Concurrent       & 4.0 days            & 7.22             \\
                                                 \multicolumn{3}{c}{} \\
                                             \end{tabular}
        \end{center}}\end{minipage}
    }
    \\
    \centering
    \vspace{.8em}
    \subfloat[
        \textbf{Strategy transferability}.\! The strategies from smaller models (\emph{Transfer}) exhibit competitive performance on larger models.\label{tab:transferability}
    ]{
        \begin{minipage}{0.42\linewidth}{\begin{center}
                                            \tablestyle{8pt}{1.15}
                                            \begin{tabular}{ccll}
                                                Model                  & Strategy& Steps$=$4                               & Steps$=$8                               \\\shline
                                                \multirow{3}{*}{Base}  & \gc{Baseline} & \gc{6.56} & \gc{5.11} \\
                                                & \default{Optimized} & \default{\textbf{3.83}}                 & \default{\textbf{2.89}}                 \\
                                                & \emph{Transfer}            & 3.87 \textcolor{gray}{\tiny{($+$0.04)}} & 3.03 \textcolor{gray}{\tiny{($+$0.14)}} \\\hline
                                                \multirow{3}{*}{Large} & \gc{Baseline} & \gc{5.42} & \gc{4.67} \\
                                                & \default{Optimized} & \default{\textbf{3.26}}                 & \default{\textbf{2.68}}                 \\
                                                & \emph{Transfer}            & 3.36 \textcolor{gray}{\tiny{($+$0.10)}} & 2.82 \textcolor{gray}{\tiny{($+$0.14)}} \\
                                            \end{tabular}
        \end{center}}\end{minipage}
    }
    \hspace{2em}
    \subfloat[
        \textbf{Hyperparameter impact on generation}.
        We exclude one of the hyperparameters from our optimization set and measure its impact on FID.
        \label{tab:abl_upd_set}
    ]{
        \centering
        \begin{minipage}{0.51\linewidth}{\begin{center}
                                             \tablestyle{1.mm}{1.15}
                                             \begin{tabular}{cccc|cc}
                                                 \multirow{2}{*}{\shortstack{Re-masking \\ratios $\bm{r}$}} & \multirow{2}{*}{\shortstack{Sampling \\Temperatures $\bm{\tau_1}$}} & \multirow{2}{*}{\shortstack{Re-masking \\Temperatures $\bm{\tau_2}$}}       & \multirow{2}{*}{\shortstack{Guidance \\scales $\bm{s}$}} & \multirow{2}{*}{FID $\downarrow$}        & \multirow{2}{*}{$\Delta$} \\
                                                 & &        & &          \\\shline\rowcolor{defaultcolor}
                                                 \cmark & \cmark & \cmark & \cmark        & \textbf{4.30}          & - \\
                                                 & \cmark & \cmark & \cmark        & 4.58          & $+$0.28 \\
                                                 \cmark & & \cmark & \cmark        & 6.57          & $+$2.27 \\
                                                 \cmark & \cmark & & \cmark        & 5.68          & $+$1.38 \\
                                                 \cmark & \cmark & \cmark &        & 5.43          & $+$1.13 \\
                                             \end{tabular}
        \end{center}}\end{minipage}
    }
    \hspace{2em}
    \vskip -0.1in
    \caption{\textbf{Ablation studies} on ImageNet-256. We report FID-50K.
    If not specified, we adopt our small model trained for 300K iterations and generate images in 4 steps.
    Default settings are marked in \colorbox{defaultcolor}{gray}.
    }
    \label{tab:ablations}
    \vskip -0.1in
\end{table*}

%% file: sec/conclusion.tex
\section{Conclusion}
\label{sec:conclusion}
In this paper, we investigated the optimal design of training and inference strategies for non-autoregressive Transformers (NATs) in image synthesis. Distinct from prior works, our approach circumvents the reliance on heuristic rules and manual fine-tuning by embracing a heuristic-free, automatic approach. This was achieved by formulating the optimal configuration of training and generation strategies as a unified hyperparameter optimization problem. We further proposed to solve this problem with a tailored alternating optimization algorithm for rapid convergence.
Extensive experiments validated that our method, \Ours, significantly enhances the performance of NATs and achieves results comparable to the latest diffusion models while requiring substantially fewer computational resources.

%% file: sec/acknowledgements.tex
\section*{Acknowledgements}
\label{sec:acknowledgements}
This work is supported in part by the National Science and Technology Major Project  (2022ZD0114903) and the National Natural Science Foundation of China (42327901, 62321005).
The authors would like to sincerely thank Yang Yue for his insightful discussion.